\documentclass[letterpaper, 10 pt, conference]{ieeeconf}  %

\IEEEoverridecommandlockouts                              %
\usepackage[backend=biber,
            url=false,
            isbn=false,
            doi=false,
            backref=false,
            style=ieee,
            natbib=true,%
            mincitenames=1,
            maxcitenames=1,
            citestyle=numeric-comp,
            sorting=nyt,%
            block=none]{biblatex}
        
\renewcommand{\bibfont}{\small}
\pdfoutput=1
\usepackage{graphics}
\usepackage[pdftex]{graphicx}
\usepackage{wrapfig}
\DeclareGraphicsExtensions{.pdf,.png,.jpg}
\usepackage{epsfig}
\usepackage[font={small}]{caption}
\usepackage{subfig}
\usepackage[rightcaption]{sidecap}
\usepackage{pbox}

\usepackage{bigstrut}
\usepackage{mathtools}
\usepackage{amsmath, amssymb, amscd}
\usepackage{ wasysym } %
\usepackage{amsfonts}
\usepackage{mathptmx} %
\usepackage{gensymb} 
\usepackage{nicefrac}       %
\numberwithin{equation}{section} 

\DeclareMathAlphabet{\mathcal}{OMS}{lmsy}{m}{n}
\DeclareSymbolFont{largesymbols}{OMX}{cmex}{m}{n}
\usepackage{textcomp} %

\usepackage{algorithm} %
\usepackage[noend]{algorithmic}

\renewcommand{\algorithmiccomment}[1]{\bgroup\hfill//~#1\egroup}
\usepackage{array} %
\usepackage{tabularx}
\usepackage{multirow}
\usepackage{multicol}
\usepackage{booktabs}
\usepackage{tabulary}

\usepackage[T1]{fontenc} 
\usepackage[utf8]{inputenc}
\usepackage[english]{babel} %
\usepackage{units}
\usepackage{bm}
\usepackage{times} %
\usepackage{xspace}
\usepackage{balance} 
\usepackage{csquotes}
\usepackage{makeidx}
\usepackage{blindtext}

\let\labelindent\relax
\usepackage[inline]{enumitem}

\usepackage{ragged2e}
\usepackage{soul} %
\usepackage{subfiles} %

\usepackage[protrusion=true,expansion=true]{microtype}
\usepackage[yyyymmdd]{datetime}

\date{\protect\formatdate{1}{1}{2001}}

\usepackage{url}
\makeatletter
\g@addto@macro{\UrlBreaks}{\UrlOrds}
\makeatother
\usepackage{color}
\usepackage{hyperref}
\hypersetup{
    colorlinks=true,
    linkcolor=black,
    citecolor=black,
    filecolor=cyan,
    urlcolor=black
}

\usepackage{marginnote}
\usepackage{soul} %

\usepackage[colorinlistoftodos]{todonotes}
\newcommand{\tocite}[1]{%
\textcolor{red}{[cite:\ifthenelse{\equal{#1}{}}{}{#1}?]}
}

\newcommand{\ignore}[1]{}

\renewcommand{\baselinestretch}{1.0}



\DeclareMathOperator{\Slerp}{Slerp}
\DeclareMathOperator{\quattoangle}{quat2angle}
\DeclareMathOperator*{\argmin}{argmin}
\DeclareMathOperator*{\argmax}{argmax}

\usepackage[colorinlistoftodos]{todonotes}

\title{\LARGE \bf
Orienting Novel 3D Objects \\
Using Self-Supervised Learning of Rotation Transforms%
}

\author{Shivin Devgon, Jeffrey Ichnowski, Ashwin Balakrishna, Harry Zhang, Ken Goldberg%
\thanks{University of California, Berkeley AUTOLAB}%
}

\begin{document}

\maketitle

\begin{abstract}
Orienting objects is a critical component in the automation of many packing and assembly tasks. We present an algorithm to orient novel objects given a depth image of the object in its current and desired orientation.
We formulate a self-supervised objective for this problem and train a deep neural network to estimate the 3D rotation as parameterized by a quaternion, between these current and desired depth images. We then use the trained network in a proportional controller to re-orient objects based on the estimated rotation between the two depth images. Results suggest that in simulation we can rotate unseen objects with unknown geometries by up to 30\degree~with a median angle error of 1.47\degree~over 100 random initial/desired orientations each for 22 novel objects. Experiments on physical objects suggest that the controller can achieve a median angle error of 4.2\degree~over 10 random initial/desired orientations each for 5 objects.
\end{abstract}

\section{Introduction}
\label{sec:introduction}
Rotating novel objects to a desired orientation is required for automating many applications including inspection, assembly, packing, and manufacturing. Consider a robot in a warehouse picking a set of objects out of a heap, scanning them to determine their identity, and then reorienting them to pack them in a specific configuration. If each object is associated with a 3D geometric model, this can be used to estimate its pose and plan a specific re-orientation. However, obtaining geometric models for novel 3D objects can be time-consuming, motivating algorithms that can reliably and precisely reorient objects without prior knowledge of object geometry.

We propose an algorithm that reorients previously unseen objects with unknown geometry given a depth image of the desired object orientation. Building on prior work on relative pose estimation for 3D objects~\cite{latent-3d-keypoints, delta-pose-est}, we leverage simulation and self-supervision to train a deep neural network to estimate a 3D rotation between two depth images. %
To represent a 3D rotation, we use quaternions as they
provide several benefits for representing and learning rotations, including fast normalization, requiring only 4 real values, and smooth interpolation not subject to singularities such as gimbal lock.
With the trained deep neural network and given a depth image of a previously unseen object in its desired orientation, we implement a controller that uses the estimated 3D rotation to reorient the object without requiring knowledge of its geometry or explicit pose estimation. 
This paper makes 4 contributions:
\begin{enumerate}%
\item A self-supervised objective to train a convolutional neural network (CNN) to estimate the rotational difference of a 3D object in two different orientations using two depth images.
\item A controller that uses the CNN to reorient objects into a desired orientation without a reference 3D model or reference orientation.
\item Simulation experiments suggesting that the controller can reliably orient 22 novel objects by up to 30\degree~with a median angle error of 1.47\degree~over 100 random initial/desired orientations per object.
\item Physical experiments suggesting that the proposed controller can reorient 5 novel objects by up to 30\degree~with a median angle error of 4.2\degree~over 10 random initial/desired orientations per object.
\end{enumerate}
\newcommand{\Tk}[1]{\includegraphics[%
width=112pt,angle=180,origin=c]{figures/#1}}
\begin{figure}
    \centering
    \begin{tabular}{c@{\hspace{5pt}}c@{\hspace{5pt}}}
    \subfloat[Rendering of pose 1]{%
        \includegraphics[clip,width=112pt]{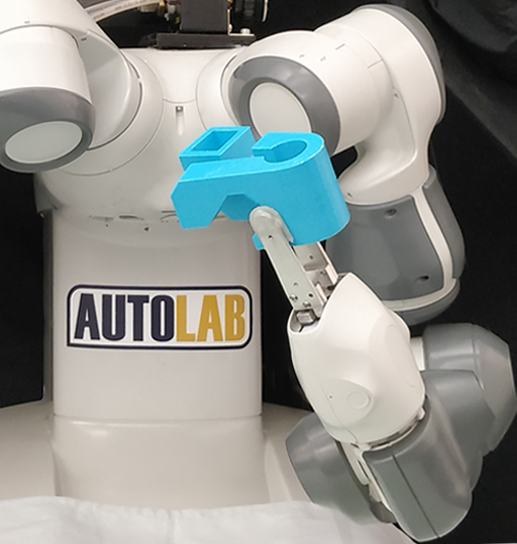}} &
    \subfloat[Rendering of pose 2]{%
        \includegraphics[clip,width=112pt]{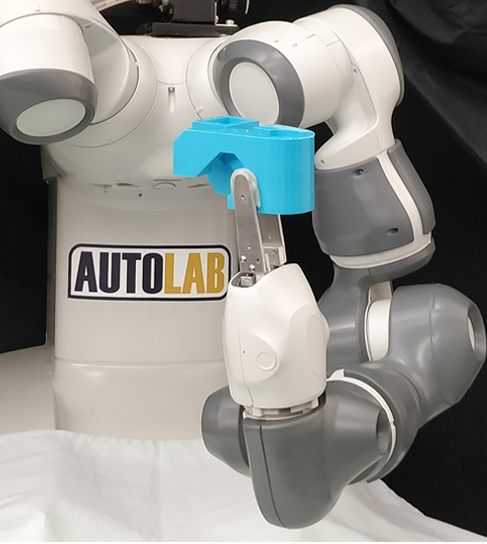}} \\
    \subfloat[Depth image of pose 1]{\Tk{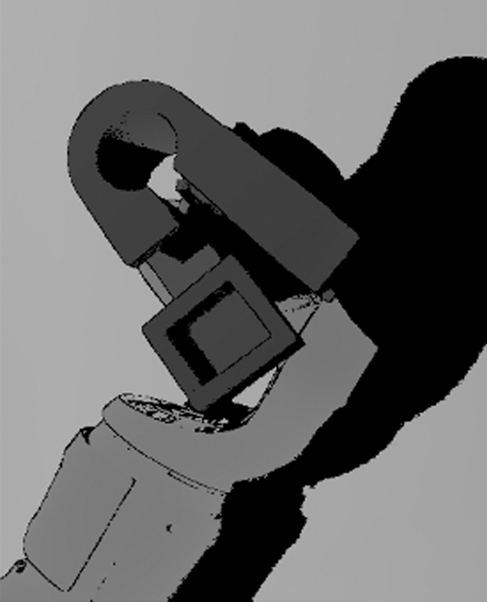}} &
    \subfloat[Depth image of pose 2]{\Tk{depth_pose2.png}}
    \end{tabular}
    \caption{\textbf{Using a depth camera, a robot gripper orients a previously-unseen object to achieve a desired pose}.
    Here the robot holds the object (a) below a depth camera.  From the depth camera's image (c), the proposed system estimates a 3D rotation to get the part into a desired orientation.  The proposed controller then re-orients the object based on the estimation to get it into the desired orientation (b).
    \textbf{Top row: }Images of a barclamp object with different poses in a gripper attached to the ABB YuMi robot. \textbf{Bottom row: } Depth images of the object viewed from the depth camera}
    \label{fig:fig1}
\end{figure}

\section{Related Work}
\label{sec:related-work}
There has been significant work in reorienting objects using geometric algorithms. \citet{goldberg1993orienting} presents a geometric algorithm to orient polygonal parts with known geometry without sensors. \citet{akella-orienting-uncertainty} extend these ideas to orienting objects with known geometry but unknown shape variations with both sensor-based and sensor-less algorithms. \citet{kumbla2018enabling} use a combination of vision and active probing to estimate an objects pose and reorient it. There has also been model-based work on reorienting objects with robot finger motions by planning grasp gaits which maintain grasp stability~\cite{grasp-gaits}. 
While all of these works require knowledge of object geometry, we propose a method which can reliably reorient objects without 3D object models. 

Another approach to reorienting objects explored by prior work uses statistical methods for pose estimation. The goal in this approach is to estimate the 6-DOF (translation + rotation) pose of an object with known geometry subject to uncertainty in sensing and occlusions. If 6-DOF poses can reliably be estimated, then re-orientation plans can be computed using the difference between the pose of the object in its initial orientation and its goal orientation. %
\citet{hodan2018bop} provides eight datasets to train and test pose estimation algorithms and a consistent benchmark that works well for evaluating various methods on symmetric and partially occluded objects. \citet{prokudin2018deep, kingma2013auto} introduce a variational-auto-encoder-based probabilistic model for pose estimation. \citet{xiang2017posecnn} use semantic labeling and bounding-box prediction as surrogate tasks to perform pose estimation via quaternion regression with a new symmetry-invariant loss function.  \citet{Li2019DeepIMDI} builds on prior work by using PoseCNN~\cite{xiang2017posecnn} to provide an initial pose estimate and then iteratively refines it by matching the image rendered based on the pose estimate and the observed image of the object. 
\citet{Do2018Deep6DPoseR6} use Mask-RCNN to perform instance segmentation and then finds a Lie algebra representation of the 6D pose of each object in a given image.
\citet{tian2020robust} learn to predict the rotation of symmetric objects by learning directly from their RGB-D features, improve upon Shape-Match Loss of \citet{xiang2017posecnn}, and include an uncertainty on the rotation prediction. 
\citet{peretroukhin2020smooth} proposes a novel representation of $SO(3)$ which incorporates the belief over the predicted rotation, making the learned model robust to unseen objects and scenes. 
\citet{hagelskjaer2019using} uses spatial reasoning and workcell constraints to accurately estimate poses. \citet{deng2019self} improve upon object segmentation and pose estimation with a self-supervised method of collecting training data from real images using an RGBD camera mounted onto the hand of a robot manipulator.
In contrast to these works, we propose a method that does not require prior knowledge of object geometry and can generalize to objects outside of those seen during training.

Some recent work explores pose estimation for objects unseen during training. \citet{Morrison2020EGAD} provide over 2000 object models for grasping and other tasks, with 49 objects specifically intended for evaluation. \citet{corona2018pose} predict the pose of objects unseen at training time, but require a 3D model to adjust for ambiguities due to symmetry. \citet{Xiao2019PoseFS} trains a pose-estimation network that is conditioned on a test image and 3D object model, making it possible to predict the pose of arbitrary objects in varied visual environments if 3D models of the objects are available. These works can generalize to unseen objects, but still require 3D object models. \citet{park2019latentfusion} relax this assumption by estimating a 3D geometric model by learning a 3D object representation that enforces consistency across multiple views. Then, this estimated 3D object model can be rendered as a depth image of the object in a desired pose. This enables generalization to objects with unknown geometry, but requires that multiple views of each object are available at test time. 
\citet{wang20196-pack} extract 3D keypoints from RGBD images of unseen objects for real-time pose tracking, but require that test objects be relatively similar to those seen in training.
\citet{Stevi2020LearningTA} estimate a goal object's pose to enable robots to execute a complex assembly task. The approach generalizes to unseen objects that contain a certain shape template. In contrast, we make no geometric assumptions about the test object, and present a method which can be applied towards reorienting novel objects with unknown geometries.

The most similar works to the proposed method are \citet{latent-3d-keypoints} and \citet{delta-pose-est}. \citet{latent-3d-keypoints} uses 3D keypoints to estimate the orientation difference between unseen objects of unknown geometries given RGB images in two poses and knowledge of the object category. \citet{delta-pose-est} estimates the relative pose between two cameras given RGB images from each. We build on these ideas and also train a network to estimate the relative orientation between two images of an 3D object.
However, in contrast to these works, we utilize simulated depth data during learning and utilize the learned network to define a controller to re-orient novel 3D objects.
\section{Problem Statement}
\label{sec:problem-statement}

\subsection{Problem Formulation}
\label{subsec:formulation}
Let $R^s \in SO(3)$ be the start rotational orientation of a rigid object of unknown geometry and let $R^g \in SO(3)$ be the goal rotational orientation of the same rigid object where $SO(3)$ is the special orthogonal group of all rotations in 3D Euclidean space.
Let $I^s \in \mathbb{R}^{H \times W}$ be a depth image observation of the object in $R^s$, and $I^g \in \mathbb{R}^{H \times W}$ be the observation of $R^g$. %

We do not have or define a reference rotational orientation, but instead estimate ${_s}R^g \in SO(3)$, such that a rotation of the object by ${_s}R^g$ reorients the object from $R^s$ to $R^g$.  Thus, $R^g = {_s}R^g R^s$ and ${_s}R^g = R^g(R^s)^{-1}$. The objective is two-fold: (1) compute an estimate of ${_s}R^g$, denoted ${_s}\hat{R}^g$, given only image observations $I^s$ and $I^g$ and (2) use this estimate to reorient the object to orientation $\hat{R}^g$ such that $\mathcal{L}({_s}\hat{R}^g, {_s}R^g)$ is minimized, where
$\mathcal{L}: SO(3) \times SO(3) \rightarrow \mathbb{R}$ is a distance measure between orientations. 
In this work, we assume that the rotation angle between $R^s$ and $R^g$ has magnitude at most 30\degree.
For objects with symmetries, the objective is to estimate and orient objects relative to a (symmetric) orientation that results in $I^g$.

\subsection{Background}
We use unit quaternions to represent rotations.
A quaternion $q = q_r + q_i i + q_j j + q_k k$ is an extension of complex numbers with a real component $q_r$ and 3 scaled fundamental imaginary units $i$, $j$, and $k$.
We represent $q$ using the convention of a vector $\begin{bmatrix} q_r & q_i & q_j & q_k \end{bmatrix}^T$. 
A \emph{unit} quaternion has the property that $\lVert q \rVert^2 = q_r^2 + q_i^2 + q_j^2 + q_k^2 = 1$, and can represent a rotation with properties we make use of in this work:
\begin{description}[left=0pt]
    \item[Normalization] A unnormalized or non-unit quaternion $\tilde{q}$ can be converted to a unit quaternion by dividing by its norm $\tilde{q} / \lVert \tilde{q} \rVert_2$
    \item[Angle difference] The angle of rotation $\theta$ between two quaternions $q_0$ and $q_1$ is $2\cos^{-1} \lvert \langle q_0, q_1 \rangle \rvert$
    \item[Rotation difference] The quaternion rotation between two quaternions is $q_{\mathrm{diff}} = q_0 q_1^{-1}$, where $q_1^{-1}$ is the conjugate (negated imaginary components) of $q_1$
    \item[Slerp] The spherical linear interpolation or \emph{slerp} between rotations $q_0$ and $q_1$ by a scalar $t \in [0,1]$ is  $\Slerp(q_0, q_1, t) = (q_0 \sin (1-t)\theta + q_1 \sin t\theta) / \sin \theta$, where $\theta$ is the angle between the two rotations~\cite{shoemake}.
    \item[Angle of rotation] The angle of rotation of a quaternion is defined by $\quattoangle(q) = 2 \cos^{-1} q_r$
    \item[Axis of rotation] The axis of rotation of a quaternion is $\begin{bmatrix} q_i & q_j & q_k \end{bmatrix} / \sqrt{1 - q_r^2}$
    \item[Double Coverage] Quaternions double cover $SO(3)$, in that $q$ and $-q$ represent the same rotation.
\end{description}

\section{Method}
\label{sec:method}
\begin{figure}
  \centering
  \includegraphics[width=0.48\textwidth]{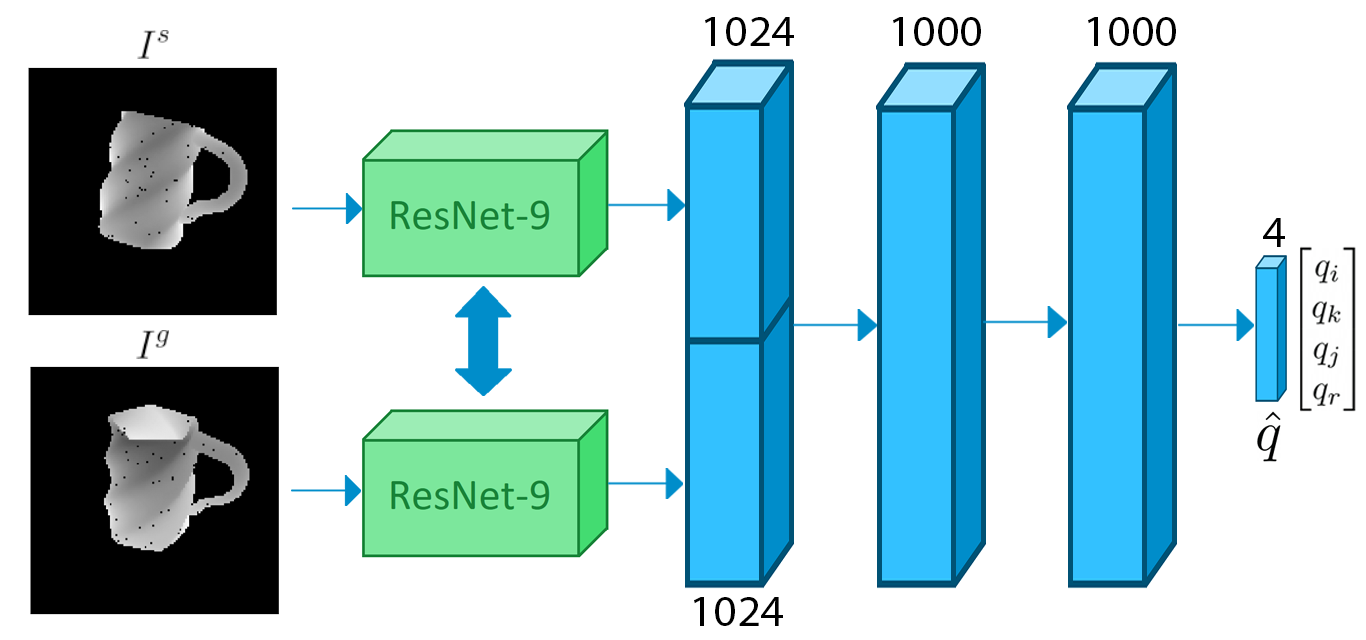}
  \caption{Given two depth images $I^s$ and $I^g$ of an object in orientations $R^s$ and $R^g$, we train a neural network to estimate the quaternion $\hat{q}$ which parameterizes the rotation difference between $R^s$ and $R^g$. Both images are fed into a ResNet-9 backbone to embed onto a feature vector of length 1024 per image. These embeddings are then concatenated and fed through two fully connected layers with Leaky ReLU activation functions, to then output a predicted $\hat{q}$.}
  \label{fig:network}
\end{figure}

Here we describe a method to train a network to compute ${_s}\hat{R}^g$ from an image pair $(I^s, I^g)$ (Section \ref{subsec:learning}).  We then describe a method that uses the trained network to create a controller for re-orienting objects using image pairs (Section \ref{subsec:reorientation-controller}). 

\subsection{Learning to Estimate 3D Rotations}
\label{subsec:learning}
To compute ${_s}\hat{R}^g$ from depth images $I^s$ and $I^g$, we use a quaternion representation for ${_s}\hat{R}^g$, denoted $\hat{q}$. We train a network $f_\theta(I^s, I^g) = \hat{q}$ to minimize some loss $\mathcal{L}(q, \hat{q})$ between $\hat{q}$ and the quaternion corresponding to the ground truth 3D rotation ${_s}R^g$ denoted by $q$, over
training dataset $\mathcal{D}_{\text{train}} = \{I^s_i, I^g_i, q_i\}_{i=1}^{N}$. Each of the $N$ datapoints in $\mathcal{D}_{\text{train}}$ is  a tuple which is composed of a pair of images and the quaternion corresponding to their relative orientation.

We consider two options for the loss function: $\mathcal{L}_{\text{mean}}$ and $\mathcal{L}_{\text{SM}}$.
$\mathcal{L}_{\text{mean}}$ is a loss function that computes the mean angle between $q$ and $\hat{q}$.
$\mathcal{L}_{\text{SM}}$ is a modification of the ShapeMatch-Loss from~\cite{xiang2017posecnn}.

The mean angle between $q$ and $\hat{q}$ averaged over the training dataset can be computed as follows:
\begin{align}
    \label{eq:mean-loss}
    \mathcal{L}_{\text{mean}}(\mathcal{D}_{\text{train}}, \theta) =\frac{1}{N}\sum_{i=1}^N \cos^{-1}(\langle q_i, f_\theta(I^s_i, I^g_i) \rangle)
\end{align}
This loss function, while effective for some objects, does not encourage invariance to symmetry. Thus, it can penalize correct predictions for objects with axes of symmetry. To avoid this, we also consider the ShapeMatch-Loss~\cite{xiang2017posecnn}, which measures the norm difference between the point clouds resulting from applying $q$ and $\hat{q}$ respectively to each point $x$ in $\mathcal{M}_i$, the vertices of the 3D object mesh for the object corresponding to training datapoint $i$.
\begin{align}
\begin{split}
    &\mathcal{L}_{\text{SM}}(\mathcal{D}_{\text{train}}, \theta) = \\
    &\frac{1}{N}\sum_{i=1}^N \frac{1}{2|{\mathcal{M}_i}|}\sum_{x_1 \in \mathcal{M}_i} \min_{x_2 \in \mathcal{M}_i} ||R(f_\theta(I^s_i, I^g_i))x_1 - R(q_i)x_2||^2_2,
\end{split}
\end{align}
where $R(q)$ is the 3D rotation matrix corresponding to quaternion $q$.

While the ShapeMatch-Loss, $\mathcal{L}_{\text{SM}}$ is designed to handle symmetries in object geometry unlike $\mathcal{L}_{\text{mean}}$, in experiments we observe poor performance when using $\mathcal{L}_{\text{SM}}$ on datasets both with and without symmetric objects while $\mathcal{L}_{\text{mean}}$ achieves better performance. The reason for this appears to be due to the double-coverage property of quaternions in which $q$ and $-q$ correspond to the same rotation. $\mathcal{L}_{\text{SM}}$ has the same loss for quaternions with opposite signs, leading it to average these results and predict quaternions with components close to 0. However, while $\mathcal{L}_{\text{mean}}$ is unable to handle symmetries, it encourages $q_r$ to be positive and thus removes the ambiguity due to the double coverage problem. To resolve this ambiguity while enforcing invariance to symmetry, we propose a hybrid loss $\mathcal{L}_{\text{hybrid}}$, which utilizes $\mathcal{L}_{\text{mean}}$ for the first epoch of learning to encourage the predicted quaternion to have positive real component and then switches to $\mathcal{L}_{\text{SM}}$ to improve invariance to symmetry. These properties allow $\mathcal{L}_{\text{hybrid}}$ to perform better in practice than either $\mathcal{L}_{\text{mean}}$ or $\mathcal{L}_{\text{SM}}$ in isolation.

\subsection{Controller}
\label{subsec:reorientation-controller}

\begin{algorithm}[t]
\caption{Reorientation Controller}
\label{alg:controller}
\begin{algorithmic}[1]
\REQUIRE Angle error threshold ($\delta$), trained $f_\theta$, step size $\eta \in (0,1]$, target image $I^g$, $K$ maximum iterations
\STATE $q^{(0)} \leftarrow $ current orientation of gripper
\FORALL{$k = 1, \ldots, K$}
    \STATE $I^{(k)} \leftarrow$ capture image
    \STATE $\hat{q} = f_\theta(I^{(k)}, I^g)$ \COMMENT{predict rotation}
    \IF{$\quattoangle(\hat{q}) \leq \delta$}
        \RETURN \COMMENT{small predicted angle, done.}
    \ENDIF
    \STATE $q^{(k)} \leftarrow \Slerp(q^{(k-1)}, q^{(k-1)} \hat{q}, \eta)$
    \STATE rotate gripper to $q^{(k)}$
\ENDFOR
\end{algorithmic}
\end{algorithm}

As predictions from the trained network empirically have errors proportional to the actual angle difference, a single rotation prediction may not reorient an object correctly.  We thus implement a proportional controller to incrementally reorient the object.   This controller is defined in Algorithm~\ref{alg:controller}.
Let $I^{(t)}$ denote an overhead image of the object at some time $t$. Given a goal image $I^g$, the controller uses $f_\theta$ to predict a rotation to align the orientation corresponding to $I^{(t)}$ ($R^t$) with that corresponding to $I^g$ ($R^g$).  In each iteration, it rotates the object in the direction of the prediction by a tunable step-size parameter $\eta \in (0,1]$, and stops once the predicted angle is small or it reaches an iteration limit. %

\section{Implementation}
\label{sec:implementation}
In this section we describe an implementation of the proposed method. We use a simulation environment (Section~\ref{subsec:sim}) to generate a dataset (Section~\ref{subsec:datagen}) which we then use to train a network (Section~\ref{subsec:train-details}).

\subsection{Simulator}
\label{subsec:sim}
To generate the dataset we use the simulation environment from \citet{danielczuk2019segmenting}.
This environment matches the target domain and application, and makes it easy to import, render, and manipulate 3D object meshes. The proposed dataset generation method samples training examples using two distributions: a state distribution, $p(R)$, which randomizes over a diverse set of object poses, and an observation distribution, $p(I|R)$, that models sensor operation and noise. To sample a single datapoint, we first sample a state defined by $(R^s_i, {_s}R^g_i) \sim p(R)$ using a dataset of 3D CAD models and randomize object orientations and possible occlusions. Next, we render synthetic depth images $I^s_i \sim p(I^s_i|R^s_i)$ and $I^g_i \sim p(I^g_i|{_s}R^g_i,R^s_i)$. Each image has resolution 128x128 pixels and is quantized to 16-bits. We implement the controller by rendering an initial depth image $I^s$ and desired depth image $I^g$ whose relative rotation angle is at most 30 \degree. Then, we iteratively rotate the object from $I^s$ towards $I^g$ using Algorithm~\ref{alg:controller} with parameters $K=100, \eta = 0.2, \delta = 0.5$.

\subsection{Dataset Generation}
\label{subsec:datagen}
\begin{figure}
  \centering
  \includegraphics[width=0.48\textwidth]{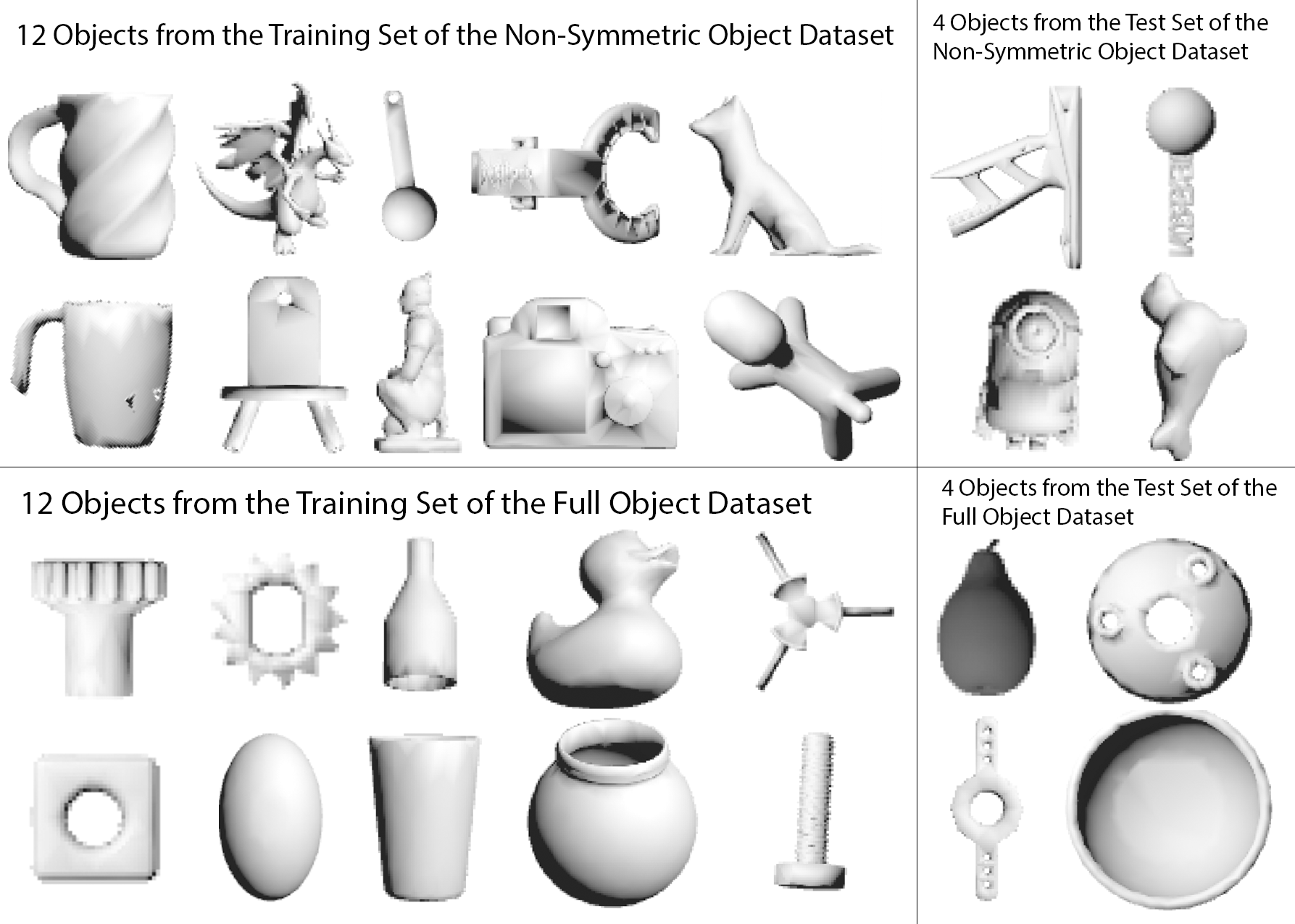}
  \caption{RGB Images of 3D objects in the non-symmetric dataset (top) and full dataset (bottom). The objects in the non-symmetric dataset  are selected based on a scoring criteria discussed in Section~\ref{subsubsec:non-symmetric}. The objects in the full dataset are taken from \cite{mahler2019learning}, and the dataset includes objects with clear symmetries.}
  \label{fig:20figs}
\end{figure}

We generate $\mathcal{D}_{\text{train}} = \{I^s_i, I^g_i, q_i\}_{i=1}^{N}$ by leveraging the same object dataset from \cite{mahler2019learning}. To generate $\mathcal{D}_{\text{train}}$, we repeatedly sample an object $O_i \in \mathcal{D}$ with replacement and do the following:
\begin{enumerate}
    \item Randomize initial orientation of $O_i$ to some $R^s_i$ which is sampled uniformly from SO3, and obtain rendered synthetic depth image $I^s_i$
    \item Apply a random 3D rotation between 0 and 30 \degree, parameterized by quaternion $q_i$ to $O_i$ resulting in a new orientation $R^g_i$ and rendered synthetic depth image $I^g_i$
    \item Store tuple $(I^s_i, I^g_i, q_i)$
\end{enumerate}
During dataset generation, we enforce that all sampled quaternions satisfy the following properties: (1) have unit-norm, (2) positive real component (if $q_r < 0$, we use $-q$ instead), (3) the magnitude of the real component must be larger than that of any imaginary component, and (4) the magnitude of the real part is at least $\cos \frac{\pi}{12}$. This process helps alleviate ambiguity due to the double-coverage property of quaternions. Note that this restricts sampling to rotation angles of magnitude at most  $\frac{\pi}{6}$ radians, since $\quattoangle(q) \le \frac{\pi}{6}$, thus $q_r \ge \cos \frac{\pi}{12}$.

Additionally, we perform domain randomization by picking a point in start image $I^s$ and generating a random thin rectangle centered at that point. This rectangle has a pixel value of zero, and so does the background. We do this because in physical experiments the object will be partially occluded by the gripper holding it. We also zero out random pixels to simulate real images taken by a Photoneo Phoxi depth camera.

\subsection{Training Details}
\label{subsec:train-details}
We use the dataset to train the network shown in
Figure~\ref{fig:network}. The two images $I^s, I^g$ first go through a ResNet-9 backbone~\cite{resnet} that embeds each image onto a feature vector. There are 5 convolutional layers, each followed by Batch-Normalization. Of these, 1 convolutional and Batch-Norm layer is used for the identity addition. Then, we flatten the channels into a size 4096 vector, go through a fully connected layer to embed each image onto a size 1024 feature vector. The two resulting embeddings are concatenated, then go through two fully connected layers of size 1000, Leaky ReLU with slope 0.02, and dropout of 0.4, a fully connected layer to regress to a size 4 vector, and finally a normalization layer to get a predicted unit-norm quaternion. The network is trained with the Adam optimizer with learning rate 0.002, decaying by a factor of 0.9 every 5 epochs with an l2 regularization penalty of $10^{-9}$. 

When training with the hybrid loss, we use vertices from the mesh of the object. For objects with over 500 vertices in the mesh, we sample 500 vertices in order to speed up training and reduce memory requirements. When using the mean angle loss defined in~\eqref{eq:mean-loss}, one issue is that numerical differentiation of the arc-cosine function can be slow and numerically unstable. Thus, to mitigate this issue, we derive an equivalent loss function as follows:
\begin{align*}
    \argmin_{\theta}\; & \mathcal{L}_{\text{mean}}(\mathcal{D}_{\text{train}}, \theta) \\
    &\equiv \argmin_{\theta}\frac{1}{N}\sum_{i=1}^N \cos^{-1}(\langle q_i, f_\theta(I^s_i, I^g_i) \rangle) \\
    &\equiv \argmax_{\theta} \sum_{i=1}^N \langle q_i, f_\theta(I^s_i, I^g_i) \rangle\\
    &\equiv \argmin_{\theta} \sum_{i=1}^N (1 - \langle q_i, f_\theta(I^s_i, I^g_i) \rangle)
\end{align*}
\section{Experiments}
\label{sec:experiments}
Here we evaluate the performance of the trained network and controller. We first evaluate the ability of the trained network from Section~\ref{subsec:learning} to predict various rotations for 3D objects both without symmetries (Section~\ref{subsubsec:non-symmetric}) and with unconstrained symmetries (Section~\ref{subsubsec:symmetric}). Then, in Section~\ref{subsec:controller-exps}, we evaluate the performance of the controller from Section~\ref{subsec:reorientation-controller} on re-orienting novel objects with unknown geometry given a single goal depth image.
\begin{figure}
  \centering
  \includegraphics[width=0.43\textwidth]{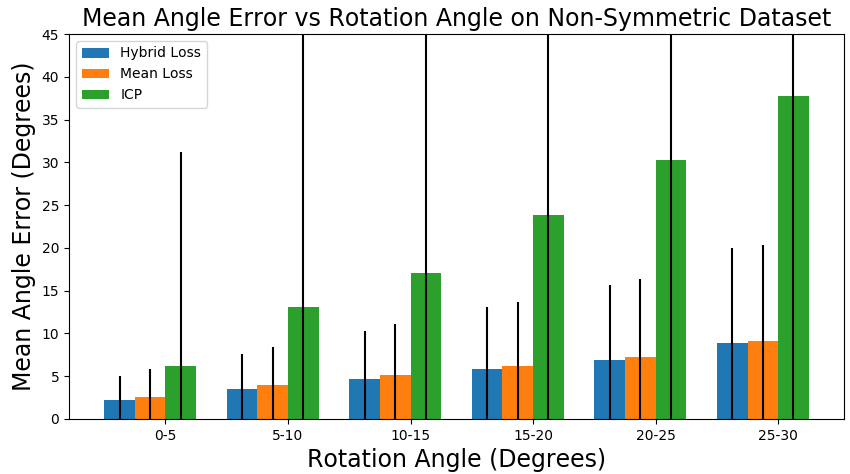}
  \caption{
  Mean angle error between $\hat{q}$ and $q$ as a function of the angle of the rotation applied by $q$ across all predictions of unseen objects in the dataset. We observe a mean angle error of 6.08\degree~when using the mean loss and a mean angle error of 5.76\degree~when using the hybrid loss. As the angle of rotation increases, so does the average error, but the ratio between the angle error and the angle of rotation decreases slightly. Results of the network trained solely on $\mathcal{L}_{\text{SM}}$ are not included because the network does not learn to adapt to the quaternion double-coverage problem. An ICP baseline is also shown.}
  \label{fig:angleloss}
\end{figure}

\begin{figure}[htbp]
  \centering
  \includegraphics[width=0.48\textwidth]{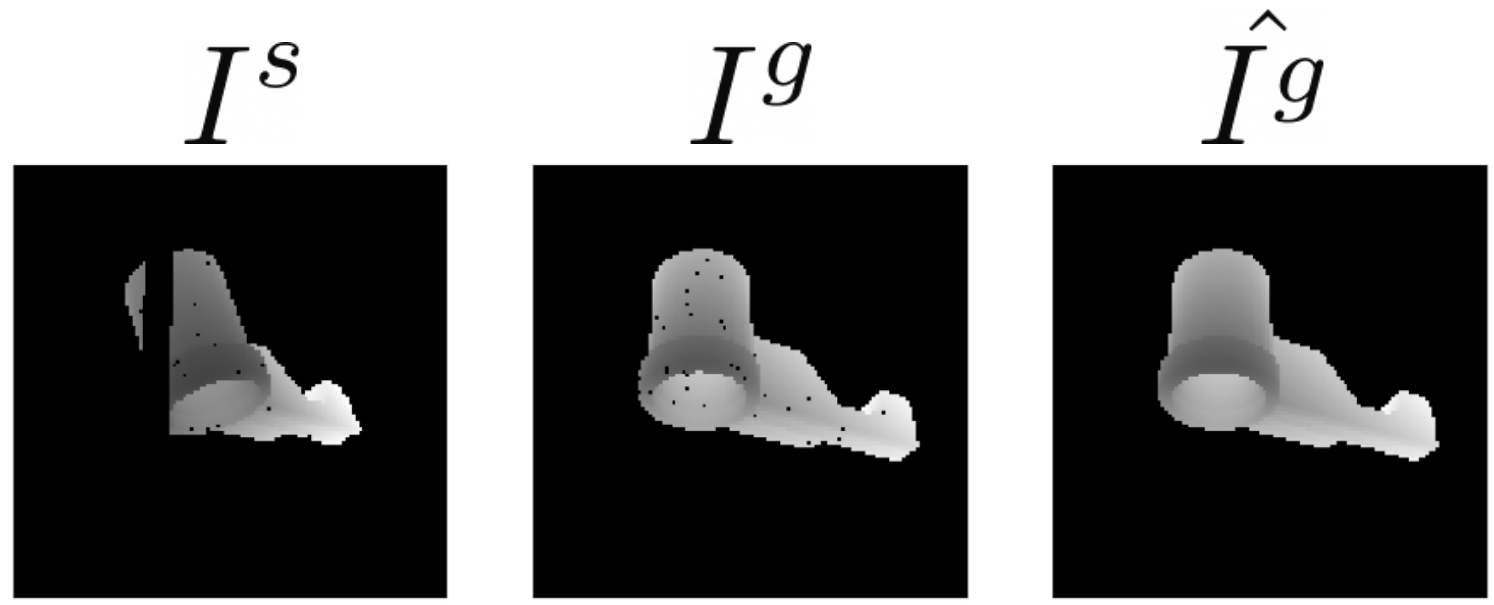}
  \caption{The best prediction of the network trained with a cosine loss on the non-symmetric dataset. The left two images are $I^s$ and $I^g$, and the right image is the result of applying $f_\theta(I^s,I^g)$ to $R^s$. $R^s$ and $R^g$ differ by a rotation of 26.36\degree.  The predicted rotation has an angle error under 0.1\degree.}
  \label{fig:best-non-symmetric}
\end{figure}

\begin{figure}[htbp]
  \centering
  \includegraphics[width=0.48\textwidth]{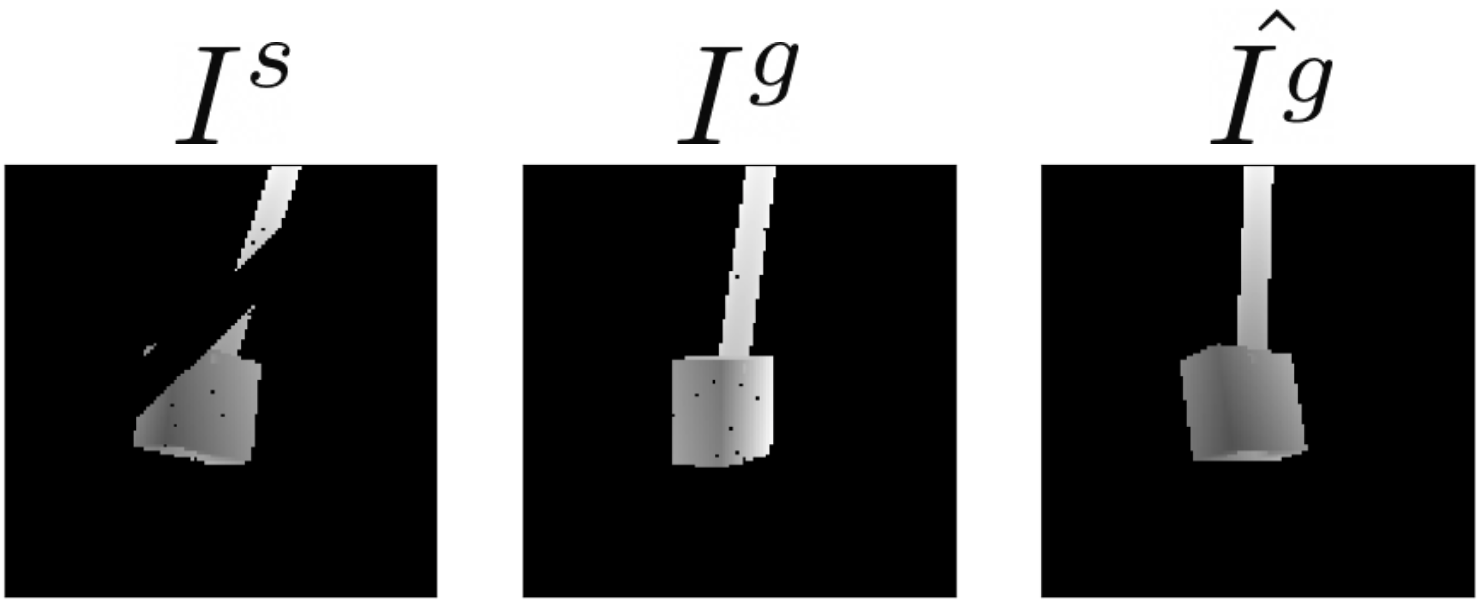}
  \caption{The worst prediction of the model of the network trained with cosine loss on the non-symmetric dataset.  The left two images are $I^s$ and $I^g$, and the right image is the result of applying $f_\theta(I^s,I^g)$ to $R^s$. $R^s$ and $R^g$ differ by a rotation of 22.59\degree. The predicted rotation has an angle error of 25.09\degree.}
  \label{fig:worst-non-symmetric}
\end{figure}

\begin{figure}[htbp]
  \centering
  \includegraphics[width=0.48\textwidth]{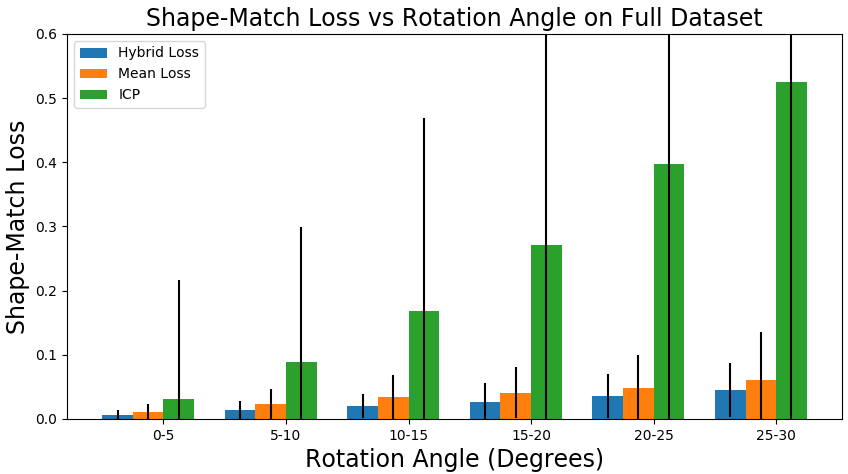}
  \caption{
  ShapeMatch-Loss between $\hat{q}$ and $q$ as a function of the angle of rotation applied by $q$ across all predictions of unseen objects in the Full dataset. As the angle of rotation increases, so does the average loss, but the ratio between the error and the angle of rotation decreases slightly. The proposed method also significantly outperforms an ICP based baseline.}
  \label{fig:symangleloss}
\end{figure}

\begin{figure*}[tb!]
    \centering
    \includegraphics[width=0.8\textwidth]{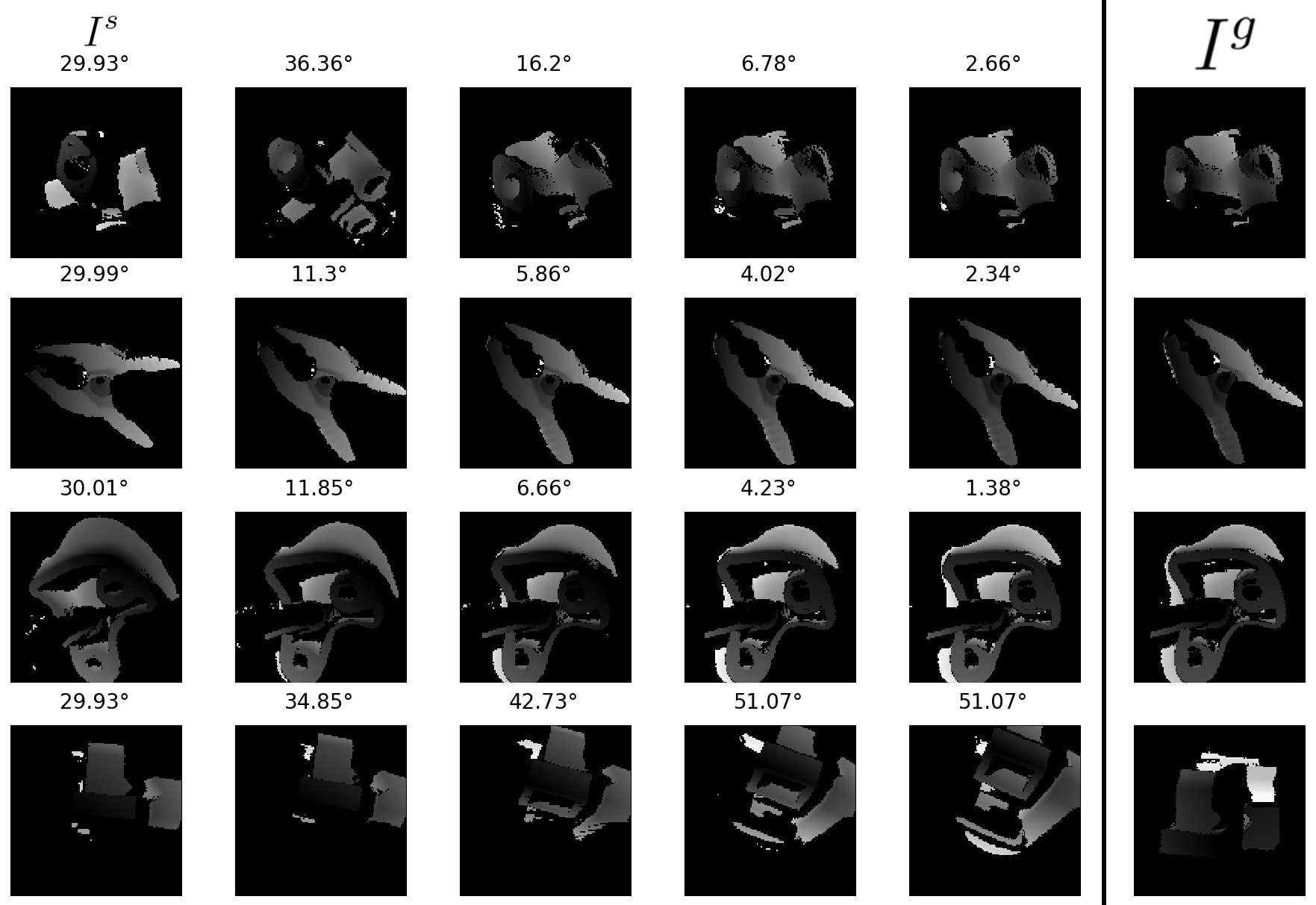}
  \caption{Visualization of controller performance of 10 trials each on $5$ physical objects. We get $I^s, I^g$ using depth and color segmentation to remove the background and gripper. The images show $I^s, I^g$, and the orientation of the objects at intermediate iterations. Over upto 50 iterations, the controller is able to accurately reorient the objects in the top three rows, but performs poorly in the bottom row.}
  \label{fig:physical_controller_evolution_images}
\end{figure*}

\label{subsec:rotation-pred}

\subsection{Predicting Rotation Transforms}
We evaluate the prediction error of the trained network on two datasets: one in which objects with clear axes of symmetries are pruned out and one with a diverse array of objects with varying degrees of symmetries. For each dataset, we present error histograms when just $\mathcal{L}_{\text{mean}}$ is used and when the hybrid loss ($\mathcal{L}_{\text{hybrid}}$) is used. Then, given the final learned network, we visualize the distribution of the angle error between the predicted quaternion and ground truth quaternion as a function of the true rotation angle applied for non-symmetric objects in Figure~\ref{fig:angleloss}. Since for symmetric objects, angle error may not always be a meaningful metric (consider predicting the rotation of a sphere, in this case any predicted angle would be correct), for the full object dataset we visualize the distribution of the ShapeMatch-Loss as a function of the true rotation angle applied for all objects in Figure~\ref{fig:symangleloss}.

\subsubsection{Non-Symmetric Object Dataset Experiments}\label{subsubsec:non-symmetric}

The first dataset contains 100 objects which are selected from the same dataset used in \cite{mahler2019learning}. Objects are selected by rejecting objects with clear axes of symmetry by (1) checking if an object mesh has an axis of symmetry detectable by the open-source Python library trimesh~\cite{trimesh}, and (2) computing all object stable poses and determining whether any of these have axes of symmetry about the x, y, or z axes. We do this by rotating each object stable pose by 120\degree~and 180\degree~around each of the x, y, and z axes, rendering the resulting point clouds before and after each rotation, and computing the point cloud distance between the initial and final point clouds. The resulting 100 objects are then randomly partitioned into a training set with $19$ objects and a test set with $81$ objects as shown in Figure~\ref{fig:20figs}. We generate a dataset using the procedure from Section~\ref{subsec:datagen} until each object has $1500$ tuples of ($I^s,I^g,q$) in the dataset. We then train the network as described in Sections~\ref{sec:method} and~\ref{sec:implementation} and evaluate its performance in Figure~\ref{fig:angleloss}. We show the error in the predicted rotation angle for different rotations for the network trained with the mean loss and the network trained with the hybrid loss and find that the angle error is significantly lower for the network trained with the hybrid loss. We also visualize the best and worst predictions of the learned network trained with hybrid loss in Figures~\ref{fig:best-non-symmetric} and~\ref{fig:worst-non-symmetric} respectively. The proposed approach also significantly outperforms a baseline which utilizes the trimesh implementation of the Iterative Closest Point (ICP) algorithm. Here we utilize the point cloud representation of $I^s$ and $I^g$ and use ICP to obtain a relative rotation and translation between the two images. Then, the relative rotation is converted to a quaternion representation for evaluation with $\mathcal{L}_{\text{mean}}$ or $\mathcal{L}_{\text{hybrid}}$.

\subsubsection{Full Object Dataset Experiments}
\label{subsubsec:symmetric}

The full object dataset contains $872$ objects which are selected from the same dataset used in \cite{mahler2019learning}. These objects are then randomly partitioned into a training set with $598$ objects and a test set with $174$ objects as shown in Figure~\ref{fig:20figs}. We generate a dataset using the procedure from Section~\ref{subsec:datagen} until each object has $200$ tuples of form ($I^s,I^g,q$) in the dataset. We then train a network as described in Sections~\ref{sec:method} and~\ref{sec:implementation} and evaluate the network's performance on the test set with respect to $\mathcal{L}_{\text{SM}}$ in Figure~\ref{fig:symangleloss}. We find that the network trained with $\mathcal{L}_{\text{hybrid}}$ performs better than the network trained with $\mathcal{L}_{\text{mean}}$. The network performs much better than an ICP baseline from the open-source Python library trimesh.

\subsection{Controller Experiments on Simulated Objects}
\label{subsec:controller-exps}
\begin{figure}
    \centering
    \includegraphics[width=0.48\textwidth]{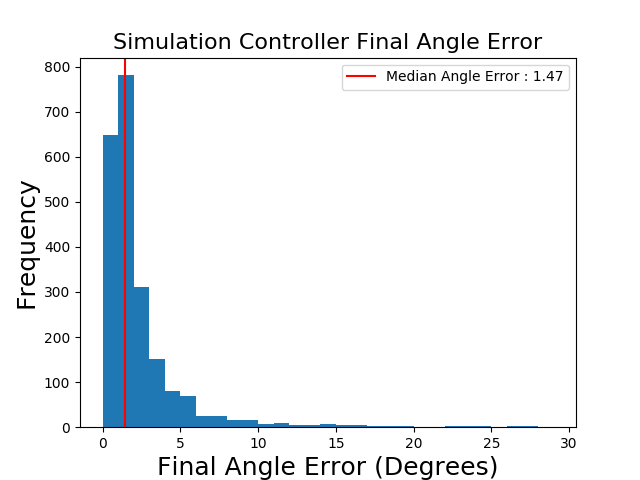}
    \caption{The final angle error between the orientation of the object at each iteration and the goal orientation for 20 randomly sampled $R^s$ and $R^g$ for 5 non-symmetric objects selected from the test set of the full object dataset. The controller is able to reorient the object within 100 iterations with a median angle error of 1.47\degree~when using $K=100, \eta=0.2, \delta=0.5 \degree$. 0.5\% of the trials had final angle errors of greater than 30\degree~(not shown).}
    \label{fig:sim_angle_hist}
\end{figure}

We implement the controller from Section~\ref{subsec:reorientation-controller} to reorient an object from start depth image $I^s$ to a desired depth image $I^g$ by using the network trained with the hybrid loss on the full dataset. For all controller experiments, we sample $I^s$ and $I^g$ such that ${_s}R^g$ corresponds to a 30\degree~rotation about an arbitrary axis. 

Figure~\ref{fig:sim_angle_hist} visualizes the distribution of the final angle errors achieved by the controller. We find that the controller is able to reorient objects successfully to within a 5\degree~angle error for 90\% of trials over 100 random initial orientations for 22 unseen non-symmetric objects. Over these same trials, we obtain a median angle error of 1.47\degree, suggesting that the controller is able to accurately reorient a variety of objects with different initial/desired orientations.

\subsection{Controller Experiments on Physical Objects}
\label{subsec:physical-exps}
\begin{figure}[h!]
    \centering
    \includegraphics[width=0.48\textwidth]{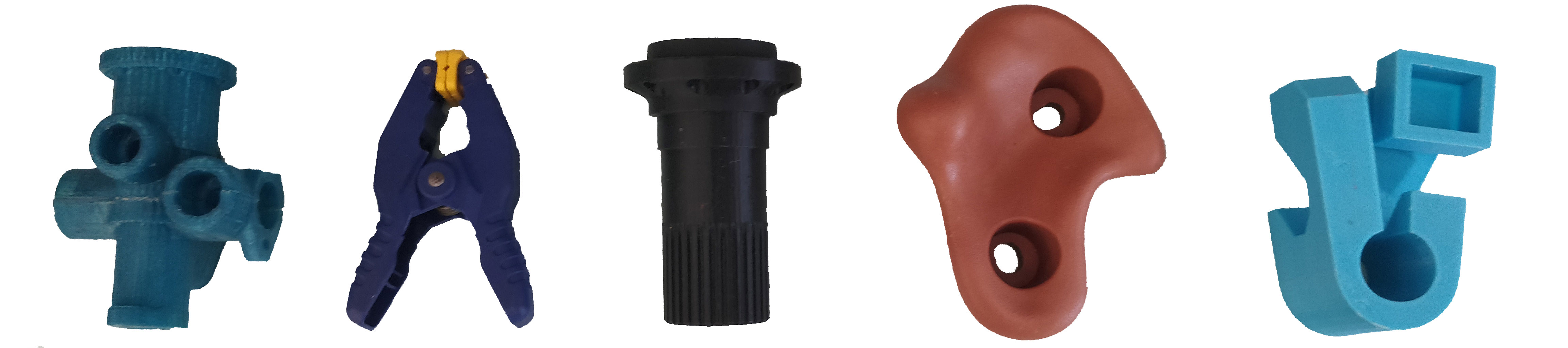}
    \caption{Real objects used in physical experiments. These objects are selected from the same object set used in~\cite{mech-search}. From the left, we used a dark blue pipe connector, a purple clamp, a black tube, a brown rock climbing hold, and a sky blue bar clamp.}
    \label{fig:real_objects}
\end{figure}

\begin{figure}[h!]
    \centering
    \includegraphics[width=0.48\textwidth]{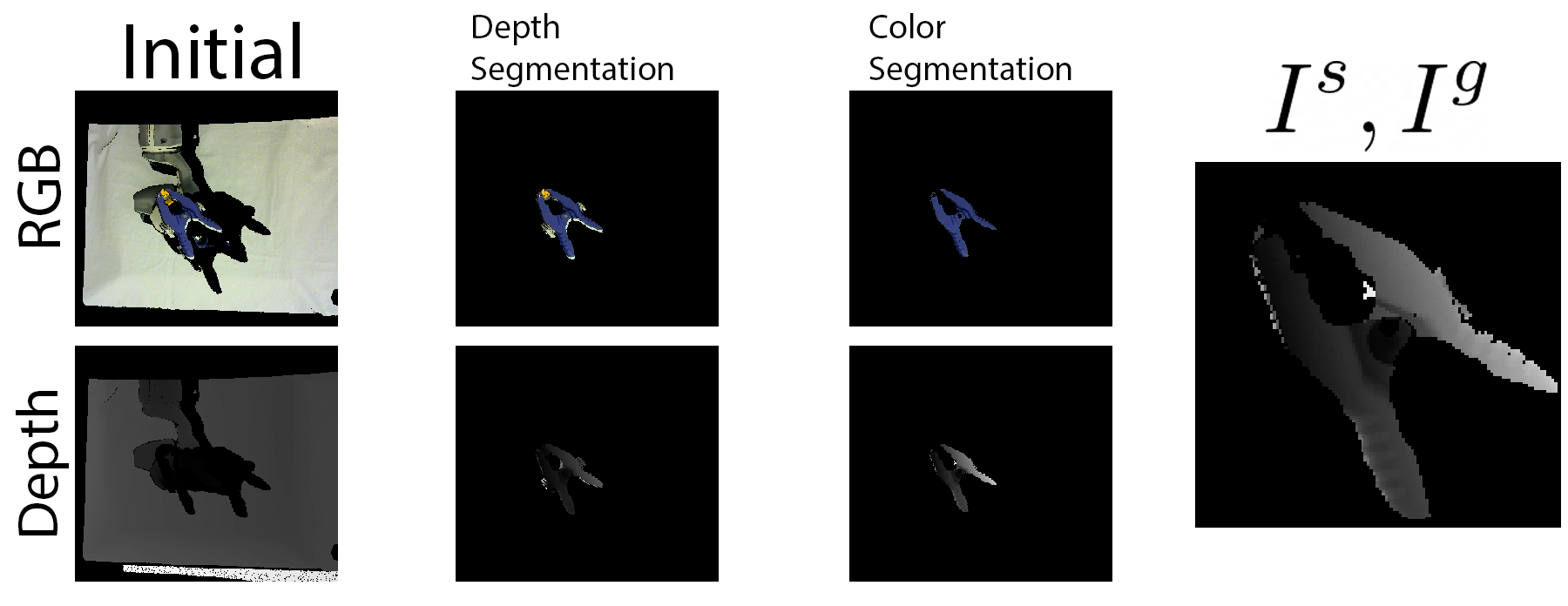}
    \caption{Physical experiments overview. First, RGB and depth images of the gripper holding the object are captured. Second, we segment the images by projecting the points to world coordinates and filtering out pixels outside the bounding box of the object. Third, we isolate the object with HSV segmentation. This generates $I^s$ and $I^g$ to feed into the neural network.}
    \label{fig:seg_pipeline}
\end{figure}

\begin{figure}[h!]
    \centering
    \includegraphics[width=0.48\textwidth]{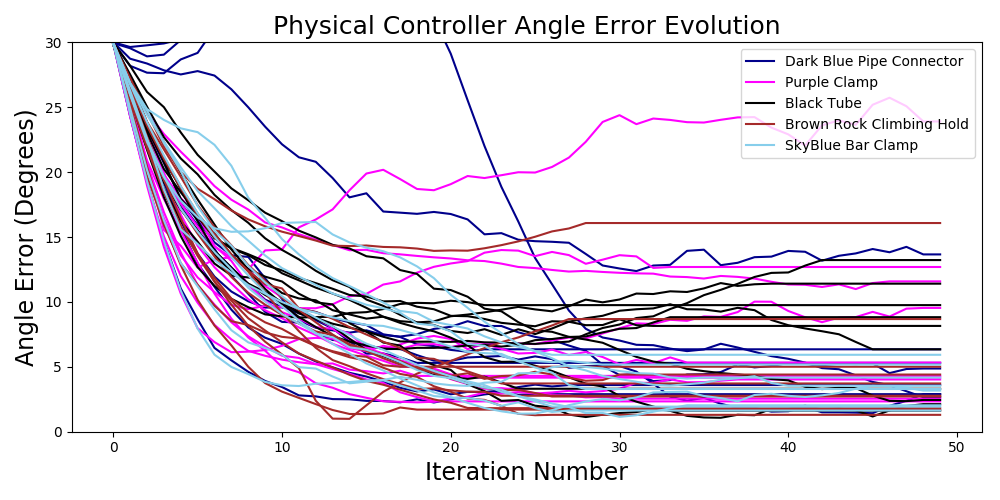}
    \caption{Angle error evolution between the orientation of the object and the goal orientation. In each iteration, the controller incrementally rotates an object to match the goal depth image. For 10 randomly sampled $R^s$ and $R^g$ for 5 physical objects the angle error is reduced with a median angle error of 4.2\degree~ over 50 trials.}
    \label{fig:physical_contoller_evolution}
\end{figure}

We evaluate the proposed method on an ABB YuMi robot with a Photoneo Phoxi camera to reorient a novel object to match the orientation in a target depth image $I^g$. We place the object inside the gripper of the YuMi and  move the gripper into a random orientation for $R^g$ such that the arm and gripper does not significantly occlude the object. We also position the object relative to the camera similarly to the objects positioned in the simulation. As illustrated in Fig. \ref{fig:seg_pipeline}, we first take an RGB image from a Logitech camera, depth image from the Photoneo Phoxi camera, and then align the RGB image into the coordinate system of the depth camera. Second, we project the depth image into world coordinates and segment out any pixels that are not within a bounding box around the object. Third, we isolate the object with HSV segmentation. Fourth, we crop and resize the image to 128x128 to generate $I^g$. We then rotate the object by 30\degree~about $({_s}R^g)^{-1}$ and repeat the process to generate $I^s$. We measure error using ground truth quaternions from the robot's forward kinematics. We then apply the controller from Section~\ref{sec:method} to reorient the object back to $R^g$. 

We evaluate the performance of the resulting controller over 10 random initial/desired orientations each for 5 physical objects shown in Figure~\ref{fig:real_objects}, where each object is tested with 10 random orientations. We use parameters $K=50, \eta = 0.2, \delta=0.5\degree$. Figure~\ref{fig:physical_contoller_evolution} shows the angle error between the image at iteration $k$ ($I^{(k)}$) and $I^g$ for these 50 trials. We find that the controller is able to successfully reduce the angle error over iterations, resulting in a final image $I^{(K)}$ which is close to the desired image $I^g$. The controller achieves a median final angle error of 4.2\degree~over these 50 trials. We notice that there is some variance in the controller performance across different start and end orientations for each given object. Thus, estimating the relative orientation between certain start and end images may be more difficult than others due to occlusions from the gripper or fewer overlapping features between the start and end orientations.
\section{Discussion and Conclusion}
\label{sec:discussion}
We present an algorithm that uses self-supervised rotation estimation network to orient novel 3D objects. Results suggest that we can orient unseen objects with unknown geometries by an initial error of up to 30\degree~with a median angle error of 1.47\degree~over 100 random initial/desired orientations each for 22 novel objects. We do this by training a deep neural network with a Resnet-9 backbone to predict the 3D rotation between an object in different initial and final orientations and use this network to define a controller to reorient an object with previously unknown geometry. Experiments on physical objects show that the controller can achieve a median angle error of 4.2\degree~over 10 random initial/desired orientations each for 5 physical objects. In future work, we will explore techniques to reduce the rotation prediction error for rotations with larger angles.
\section{Acknowledgments}
\begin{scriptsize}
\noindent This research was performed at the AUTOLAB at UC Berkeley in affiliation with the Berkeley AI Research (BAIR) Lab, Berkeley Deep Drive (BDD), the Real-Time Intelligent Secure Execution (RISE) Lab, and the CITRIS "People and Robots" (CPAR) Initiative. Authors were also supported by the Scalable Collaborative Human-Robot Learning (SCHooL) Project, a NSF National Robotics Initiative Award 1734633, and in part by donations from Siemens, Google, Toyota Research Institute, Autodesk, Honda, Intel, Hewlett-Packard and by equipment grants from PhotoNeo and Nvidia. Ashwin Balakrishna is supported by an NSF GRFP. This article solely reflects the opinions and conclusions of its authors and do not reflect the views of the sponsors. We thank our colleagues who provided feedback and suggestions, in particular Mike Danielczuk and Kate Sanders.
\end{scriptsize}

\renewcommand*{\bibfont}{\footnotesize}
\printbibliography %

\end{document}